\documentclass[11pt,a4paper]{article}
\PassOptionsToPackage{hyphens}{url}
\usepackage{natbib}
\usepackage[hyperref]{acl2021}
\usepackage{times}
\usepackage{latexsym}

\usepackage{graphicx}

\usepackage{microtype}

\aclfinalcopy 

\title{A Survey of NLP-Related Crowdsourcing HITs: what works and what does not}

\author{Jessica Huynh \\
  Carnegie Mellon University \\
  \texttt{jhuynh@cs.cmu.edu} \\\And
  Jeffrey Bigham \\
  Carnegie Mellon University \\
  \texttt{jbigham@cmu.edu} \\\And
  Maxine Eskenazi \\
  Carnegie Mellon University \\
  \texttt{max@cmu.edu} \\}

\date{}

\begin{document}
\maketitle
\begin{abstract}
Crowdsourcing requesters on Amazon Mechanical Turk (AMT) have raised questions about the reliability of the workers. The AMT workforce is very diverse and it is not possible to make blanket assumptions about them as a group. Some requesters now reject work en mass when they do not get the results they expect. This has the effect of giving each worker (good or bad) a lower Human Intelligence Task (HIT) approval score, which is unfair to the good workers. It also has the effect of giving the requester a bad reputation on the workers' forums. Some of the issues causing the mass rejections stem from the requesters not taking the time to create a well-formed task with complete instructions and/or not paying a fair wage. To explore this assumption, this paper describes a study that looks at the crowdsourcing HITs on AMT that were available over a given span of time and records information about those HITs. This study also records information from a crowdsourcing forum on the worker perspective on both those HITs and on their corresponding requesters. Results reveal issues in worker payment and presentation issues such as missing instructions or HITs that are not doable.
\end{abstract}

\section{Introduction}
\label{sec:introduction}
The rise of artificial intelligence has led to a surge in the need for data. Models are becoming increasingly more complex, needing more fine-grained data. In order to quickly collect such data, many researchers have turned to crowdsourcing. The crowdsourcing platform that is most familiar to requesters is Amazon Mechanical Turk (AMT). Anyone in many countries can create an Amazon account and offer work, in the form of HITs, to AMT workers. Desiring rapid results, requesters often post HITs quickly, not taking into account that the way in which a HIT is presented has a direct effect on the quality of the data they obtain. They may also not be aware that their actions as requesters affect the quality of the data they gather. For example, a HIT could have unclear instructions, low payment, or in some cases might not even function. The two former cases result can lead to lower data quality. Lately, this has lead requesters, seeing how to revise their HIT for better results, to refuse to pay for their earlier errorful HIT. Thus they send out mass rejections (rejecting all of the work on a given HIT regardless of its quality). While for the requester this is just restarting a HIT, the effect of this mass rejection is felt on both sides. The workers are not paid for the time they spent regardless of the quality of their work. Also, since their work has been rejected, their individual rating goes down, thus making it hard for them to qualify in the future for other HITs. The requester also gets a bad reputation on the quality of their HITs and their payments on the workers' forums such as Turkopticon \cite{irani2013turkopticon} \footnote{\url{https://turkopticon.net/}} \footnote{\url{https://turkopticon.info/}}, TurkerView \footnote{\url{https://turkerview.com/}}, Turker Nation \footnote{\url{https://www.reddit.com/r/TurkerNation/}}, and MTurk Crowd \footnote{\url{https://mturkcrowd.com/}}. These sites are regularly visited by many of the workers to determine which HITs are considered to be a reliable source of income. They also use the information to avoid certain requesters. Requesters can monitor their reputation on these sites and use worker feedback to improve their HITs and thus eventually their reputations.

\section{Background}
\label{sec:background}

Recently, recruiting workers and bringing them into a lab has become much less desirable due to the high cost and the lack of diversity of the workers. In response to this issue AMT has grown in popularity amongst researchers \cite{paolacci2010running}. 

\subsection{Worker Payment}
A major issue affecting both the interaction between the requester and the worker and the data quality is worker payment. Hara et al \cite{hara2018data} have shown that the mean wage for a worker on AMT is very low, \$3.13 per hour at the time, while requesters generally pay on average \$11.58 per hour. This apparent disparity is due to the fact that the most abundant HITs tend to be the lower paid ones. Researchers may believe that AMT workers are willing to work for very little, and that the amount of compensation does not affect data quality \cite{buhrmester2016amazon} \cite{mason2009financial}. Indeed, a good worker will try to work on the better paying HITs when possible. But if it is the end of the month and rent is due, workers are obliged to take whatever HITs are available at that time even if they don't pay well. This may result in the worker spending less time reading instructions or in actually working on a HIT, in order to make the meager payment cover less work time. This usually results in lower quality HIT data.

\subsection{Communication}
In addition to worker payment, the quality of communication between requesters and workers is frequently mentioned on forums such as Turkopticon \cite{hanrahan2021expertise}. Good communication is important when a worker wants to tell a requester about issues they have when attempting to do a HIT or when they want to find out why their work has been rejected. Good communication can lead to better data quality since it not only provides the worker with a better idea of what the requester really wants, but also helps the requester to be aware of any issues with a given HIT.

\subsection{Rejecting Work}
Rejections are a problem for both the requester and the worker. If a requester rejects work and does not use some subset of the data they collected, they may need to re-publish their HIT so that they can fulfill the originally desired quantity of data. This delays their work on their results. For the worker, if their work is rejected, their reputation suffers and they may not meet certain qualifications like HIT approval percentage for well-paying HITs \cite{hara2018data}. Even though rejections constitute a small percentage of work \cite{hara2018data}, real workers who have honestly put effort into their work should be paid for that time. On the other hand, requesters feel justified in rejecting work that has obviously come from bots. While this seems reasonable, lately the creators of those bots have started to write requester reviews, thus tarnishing the reputation of the requester. AMT suggests that when a requester is sure that a specific worker account is a bot, they should pay them and at the same time report them to AMT.

There are several other factors that affect the quality of the work, such as HITs that do not work (buggy), HITs that have no standard for evaluation \cite{mcinnis2016taking} \cite{schmidt2016using} and poor instructions. 

The aforementioned websites and worker forums are used, by about 60\% of all workers \cite{yang2018crowdworker}. Thus it is important to be attentive to worker feedback.

\section{Approach}
\label{sec:approach}
In order to prove that miscommunication and misunderstanding that affects data quality often originates with the requesters, this paper presents a study that examines available AMT HITs over a specific period of time. The first author of this paper registered as a worker and examined the HITs that were available. It should be noted that not all HITs are available to beginning workers. There are two parts to this study: recording information about all available AMT HITs during the given time period and recording information from the TurkerView website that reviews both HITs and requesters during the same time period.

\subsection{AMT}
The study ran from August 25, 2021 to August 31, 2021. Every day during this period, at 12 PM EST, the first author refreshed the HITs page that she could see as a new worker and recorded information about the available HITs that were NLP- or speech-related (vision tasks, surveys and information retrieval from websites were not included in this study) and noted the following:
\begin{itemize}
    \item Name of the HIT
    \item Requester Name
    \item Qualifications Needed
    \item Payment
    \item Time the HIT has been up
\end{itemize}

\noindent HITs fell into three categories: 
\begin{itemize}
    \item NOQUAL - HITs the first author, as a new worker, did not qualify for and was unable to preview
    \item PREVIEW - HITs the first author did not qualify for but could preview
    \item QUAL - HITs the first author qualified for
\end{itemize}

When a HIT could be previewed, it could be evaluated for the study but it was simply not possible to submit the HIT. 

\noindent For PREVIEW and QUAL, the following were also recorded:
\begin{itemize}
    \item Technical Issues
    \begin{itemize}
        \item Inaccessible - the HIT is completely inaccessible and cannot be viewed.
        \item Preview - the HIT does not work properly in preview, but might work for the workers who are allowed to work on it. 
        \item Critical - HIT does not work properly when the first author accepts to work on it.
    \end{itemize}
    \item Instruction Issues
    \begin{itemize}
        \item Completely Unclear - Instructions and/or examples do not make sense, given the task or have grammatical errors that impede comprehension.
        \item Incomplete - Instructions and/or examples are provided for some but not all annotation/mentioned categories.
        \item Ambiguous/Vague - Instructions and/or examples are vague and open to several possible interpretations.
        \item Unrelated - Instructions and/or examples do not help in completing that specific HIT. When actually working on the HIT, the instructions (that made sense during the read-through) did not help in actually doing the task.
    \end{itemize}
    \item Other comments about the HIT itself
\end{itemize}

For examples of these categories, please refer to Appendix A.

\subsection{TurkerView}
TurkerView was used to record the workers' perspective on the requester and on their HIT. The following were recorded from TurkerView: 
\begin{itemize}
    \item Worker Assessment of Payment for the HIT
    \begin{itemize}
        \item Generous
        \item Well
        \item Fair
        \item Poor
        \item Very Badly
    \end{itemize}
    \item Worker Average Recorded Payment for the HIT (average payment/average time)
    \item Worker Assessment of Payment for the Requester
    \begin{itemize}
        \item Generous
        \item Well
        \item Fair
        \item Poor
        \item Very Badly
    \end{itemize}
    \item Worker Average Recorded Payment for the Requester
    \item Communication
    \begin{itemize}
        \item poor
        \item ok
        \item good
    \end{itemize}
    \item Approves Quickly
    \begin{itemize}
        \item slowly
        \item average
        \item quickly
    \end{itemize}
    \item Rejections
    \item Blocks
\end{itemize}

Metrics covering the requester's communication (unacceptable, poor, acceptable, good, excellent) and their approval time (very slow, slow, average, fast, very fast) were also noted.

See Appendix B for a detailed breakdown of the TurkerView metrics. 

Since this study was run over the course of one week, any bias related to a particular day of the week was eliminated. 

\section{Results}
\label{sec:results}

\subsection{Overall}
There were a total of 161 HITs with the three categories (NOQUAL, PREVIEW and QUAL) as seen in Table \ref{table:overall_stats}. Several of these HITs were batches of the same HIT that were released on different days, possibly with different pay levels. These were counted as duplicate HITs if they had the same HIT name. In total, there were 102 unique HITs. These HITs were posted by 79 different requesters. Of the total of 102 HITs, 56 unique HITs were in the PREVIEW or QUAL categories. The rest of the paper concerns this set of 56 HITs. 

\begin{table}[h!]
\centering
\begin{tabular}{|c|c|c|c|} 
 \hline
 & NOQUAL & PREVIEW & QUAL \\
 \hline\hline
 Unique & 46 & 36 & 20 \\
 \hline
 Duplicate & 51 & 2 & 6 \\
 \hline\hline
 Total & 97 & 38 & 26 \\
 \hline
\end{tabular}
\caption{HIT Statistics}
\label{table:overall_stats}
\end{table}

\subsection{Issues}

As seen in Table \ref{table:tech_issues}, a total of 14 HITs (25\% of total HITs!) had technical issues. 

\begin{table}[h!]
\centering
\begin{tabular}{|c|c|c|}
 \hline
 Technical Issue & No. HITs & \% of HITs \\
 \hline \hline
 Preview & 6 & 11\%\\
 \hline
 Inaccessible & 7 & 13\% \\
 \hline 
 Critical & 1 & 2\%\\
 \hline
\end{tabular}
\caption{Technical Issues}
\label{table:tech_issues}
\end{table}

Concerning instructions, some HITs had unclear instructions while 12 HITs had incomplete instructions. This may cause the worker to complete the HIT in an undesired way because not all the terms were defined and accompanied by examples. Both unrelated and ambiguous/vague instructions will produce the same result. There were 15 HITs (27\% of the total HITs). The total number of HITs with technical and instructional issues is 17 HITs (one HIT had both a technical and instructional issue) (30\% of the total HITs!). 

\begin{table}[h!]
\centering
\begin{tabular}{|c|c|c|} 
 \hline 
 Instr. Issue & No. HITs & \% of HITs \\
 \hline \hline
 Completely Unclear & 0 & 0\% \\
 \hline
 Incomplete & 12 & 22\% \\
 \hline
 Unrelated & 2 & 4\% \\
 \hline
 Ambiguous/Vague & 1 & 2\% \\
 \hline
\end{tabular}
\caption{Instruction Issues}
\label{table:instr_issues}
\end{table}

\subsection{Payment per HIT}
Table \ref{table:payment_amt}, shows the average dollar amounts per hour that the HITs paid based on worker feedback in TurkerView. This is for 54 out of 102 recent HIT, which may not correspond to the 56 HITs discussed above. The amounts were calculated by averaging the amount the workers were paid (including bonuses for some HITs) and dividing that by the average amount of time the workers spent on the HIT. The HITs were further broken down into three categories based on TurkerView's wage assessment, which can be seen in Appendix B. 24 HITs (44\%!) paid less than \$7.25, which is below the US Federal minimum wage, 9 (17\%) paid between \$7.25 and \$10.00, and the remaining 21 (39\%) paid above \$10. The 17 HITs (31\%) paid above \$15.00 an hour, is 17 .

\begin{table}[h!]
\centering
\begin{tabular}{|c|c|c|} 
 \hline 
 Payment & No. HITs & \% of HITs \\
 \hline \hline
 $<$ \$7.25 & 24 & 44\% \\
 \hline
 \$7.25 - \$10.00 &  9 & 17\% \\
 \hline
 $>$ \$10.00 & 21 & 39\% \\
 \hline
\end{tabular}
\caption{Payment Statistics for HITs}
\label{table:payment_amt}
\end{table}

\subsection{Overall Worker View of Requester}
In addition reviewing each HIT, workers can review the requesters themselves. Table \ref{table:payment_req} shows a total of 67 requesters out of 79 that workers had been reviewed for payment assessment. 

\begin{table}[h!]
\centering
\begin{tabular}{|c|c|c|} 
 \hline 
 Assessment of Pay & No. Reqs & \% of Reqs \\
 \hline \hline
 very badly & 11 & 16\% \\
 \hline
 poorly & 13 & 19\% \\
 \hline
 fairly & 14 & 21\% \\
 \hline 
 well & 15 & 22\% \\
 \hline
 generously & 14 & 21\% \\
 \hline
\end{tabular}
\caption{Assessment of Payment for Requesters}
\label{table:payment_req}
\end{table}

72 requesters were reviewed on the average payment per hour offered over all of their reviewed HITs, as seen in Table \ref{table:payment_all}. If the payment per hour threshold is increased to \$15.00, the number of requesters is 28, (39\% of requesters as opposed to 64\% for payments over \$10.00).

\begin{table}[h!]
\centering
\begin{tabular}{|c|c|c|} 
 \hline 
 Payment & No. Reqs & \% of Reqs \\
 \hline \hline
 $<$ \$7.25 & 18 & 25\% \\
 \hline
 \$7.25 - \$10.00 &  8 & 11\% \\
 \hline
 $>$ \$10.00 & 46 & 64\% \\
 \hline
\end{tabular}
\caption{Payment Statistics for Requesters}
\label{table:payment_all}
\end{table}

Table \ref{table:comms} shows communication scores for 19 requesters. TurkerView groups unacceptable and poor together, renames acceptable as ok, and groups good and excellent together in the overall metrics. Thus 26\% of the requesters reviewed had poor communication with the workers!

\begin{table}[h!]
\centering
\begin{tabular}{|c|c|c|} 
 \hline 
 Communication & No. Reqs & \% of Reqs \\
 \hline \hline
 poor & 5 & 26\% \\
 \hline
 okay & 1 & 5\% \\
 \hline
 good & 13 & 68\% \\
 \hline
\end{tabular}
\caption{Assessment of Communication for Requesters}
\label{table:comms}
\end{table}

The requesters' ratings were broken down into approval time, rejections, and blocking workers in Tables \ref{table:approval}, \ref{table:rejections}, \ref{table:blocks}. Again, TurkerView groups very slow and slow, and fast and very fast. FOrtunately a large majority of these requesters are quick to respond to workers.

\begin{table}[h!]
\centering
\begin{tabular}{|c|c|c|} 
 \hline 
 Approval Time & No. Reqs & \% of Reqs \\
 \hline \hline
 slowly & 5 & 9\% \\
 \hline
 average & 0 & 0\% \\
 \hline
 quickly & 50 & 91\% \\
 \hline
\end{tabular}
\caption{Assessment of Approval Time for Requesters}
\label{table:approval}
\end{table}

\begin{table}[h!]
\centering
\begin{tabular}{|c|c|c|} 
 \hline 
 Rejections & No. Reqs & \% of Reqs \\
 \hline \hline
 no & 62 & 86\% \\
 \hline
 yes & 10 & 14\% \\
 \hline
\end{tabular}
\caption{Assessment of Rejections for Requesters}
\label{table:rejections}
\end{table}

\begin{table}[h!]
\centering
\begin{tabular}{|c|c|c|} 
 \hline 
 Blocks & No. Reqs & \% of Reqs \\
 \hline \hline
 no & 71 & 99\% \\
 \hline
 yes & 1 & 1\% \\
 \hline
\end{tabular}
\caption{Assessment of Blocks for Requesters}
\label{table:blocks}
\end{table}

\section{Discussion}
\label{sec:discussion}

\subsection{HITs}
Technical issues and instruction issues play a large role in whether the requester will obtain the data that they expect. The inaccessible and critical issues keep the worker from completing the HIT. Incomplete instructions, the main instruction issue, could cause some workers to interpret the task in a way that the requesters did not expect. 30\% of all the HITs fall in this category. Consent forms are often absent from in the HITs. This could be due to differing IRB standards, or on what the data from the HIT is used for.

\subsection{Requesters}
The main point of issue that appears in the workers' reviews of the requesters is the amount of payment. Even though the requesters are distributed evenly across the different levels of pay, almost 40\% of the requesters do not pay an amount that the workers judge to be fair. And, while 64\% of the requesters pay above \$10 an hour, many of the available HITs paid less than that amount. Also, workers recorded rejections from 10 out of 72 requesters. This means that these 10 requesters rejected some amount of work, whether it be specific whole batches or work from specific workers. 3 of these were below 2\%, but there 4 requesters who were above 25\%. This means that workers did work but were not paid for it. This type of result is certainly a red flag for workers choosing which requesters to work for.

\subsection{Qualifications}
On AMT, a requester can choose from a variety of types of qualifications. There are several trends within the three categories of HITs. Requesters who restricted preview privileges for the HIT tended to have more requester-specific qualifications, for example, qualification tests, specific workers, or even prior qualifications that were manually assigned due to past good work by a specific worker. They also used the Masters qualification which requires workers to have done a certain number of HITs (usually above 500 or 1000), and requested workers to have a certain approval rating (generally over 95\%). Although the requesters who allowed the first author to preview were less stringent on qualifications, they still made frequent use of at least one of the above qualifications. Finally, the HITs that the first author was qualified for either only used a high approval rating, or other qualifications such as location (that the previous categories of requesters implemented as well) that did not greatly restrict the worker population.

\subsection{Caveats}
Several variables may cause bias in our statistics, mainly:
\begin{itemize}
    \item Accessibility of the HIT
    \item Availability of the HIT
    \item Reviews of the Requesters
\end{itemize}

\subsubsection{Accessibility}
The HITs were split into the three aforementioned categories of NOQUAL, PREVIEW and QUAL. Detailed information and evaluation about the instructions for HITs could only be retrieved from the PREVIEW and QUAL HITs. Certain requesters have pre-surveys attached to their HIT on external websites; because the HIT qualifications were not met and their database may be storing intermediate results, there is no data on these HITs.

\subsubsection{Availability}
Some HITs may be completely taken immediately by workers, possibly bots or webscrapers, thus not every HIT that was released over the week of the study may have been captured. Yet other HITs may be released and taken by workers at different times of day. Some requesters may choose to have their HITs be invisible to everyone who is not qualified for their HIT \footnote{\url{https://blog.mturk.com/launch-requesters-now-have-more-control-over-hit-visibility-via-the-mturk-api-55ffdf5e7d4}}. 

\subsubsection{Reviews on TurkerView}
TurkerView does not have information for all of the requesters of the HITs in the study. This may be due to a requester not having many HITs out for review, being a newer requester, etc. This accounts for numbers in the sections in this paper that concern reviewers that are different from the numbers concerning HITs.

\section{Conclusion}
This paper reveals that, for a significant amount of HITs, there is a disconnect between the workers and the requesters. Payment is often not fair in both the workers' eyes and according to federal wage standards. HIT instructions, communication, and understanding of rejections need improvement for requesters to express what they are actually looking to obtain and thus for them to find that their data is of the quality they were expecting. 

\section*{Acknowledgments}
This material is based upon work supported by the National Science Foundation Graduate Research Fellowship under Grant Nos. DGE1745016 and  DGE2140739. It is also partly funded by the National Science Foundation grant CNS-1512973. The opinions expressed in this paper do not necessarily reflect those of the National Science Foundation.

\bibliographystyle{acl_natbib}
\bibliography{anthology,acl2021}

\appendix
\section{Examples of Issues}
\begin{figure}[h]
\includegraphics[scale=0.25]{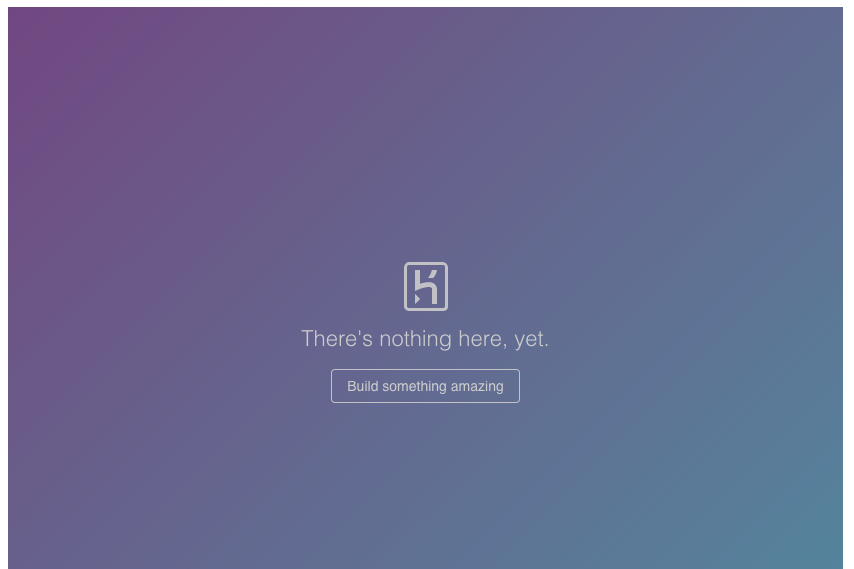}
\centering
\caption{Inaccessible Issue; Heroku error pops up on every HIT with this issue}
\end{figure}

\begin{figure}[h]
\includegraphics[scale=0.2]{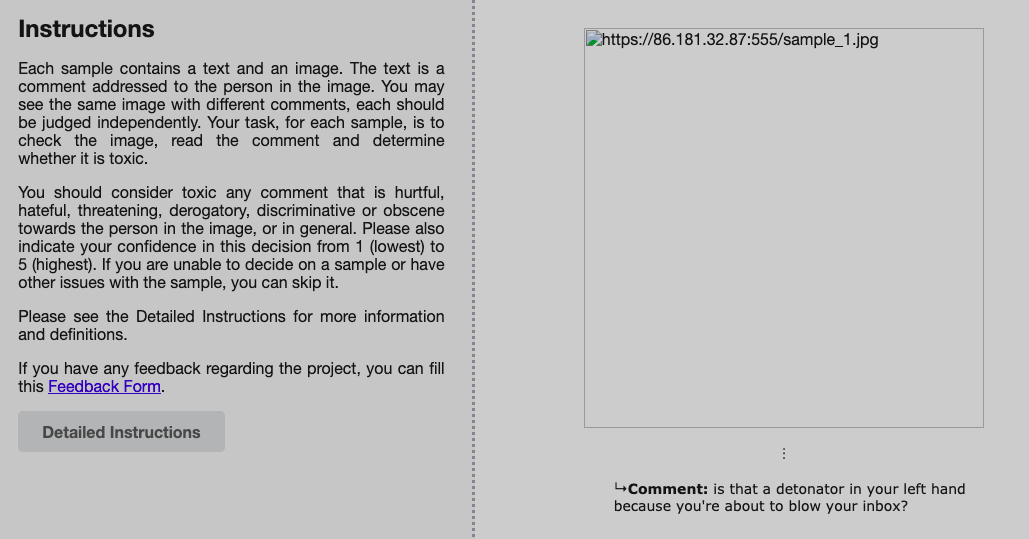}
\centering
\caption{Critical Issue; the image associated with the HIT does not show up, which prevents the worker from doing the HIT properly}
\end{figure}

\begin{figure}[h]
\includegraphics[scale=0.15]{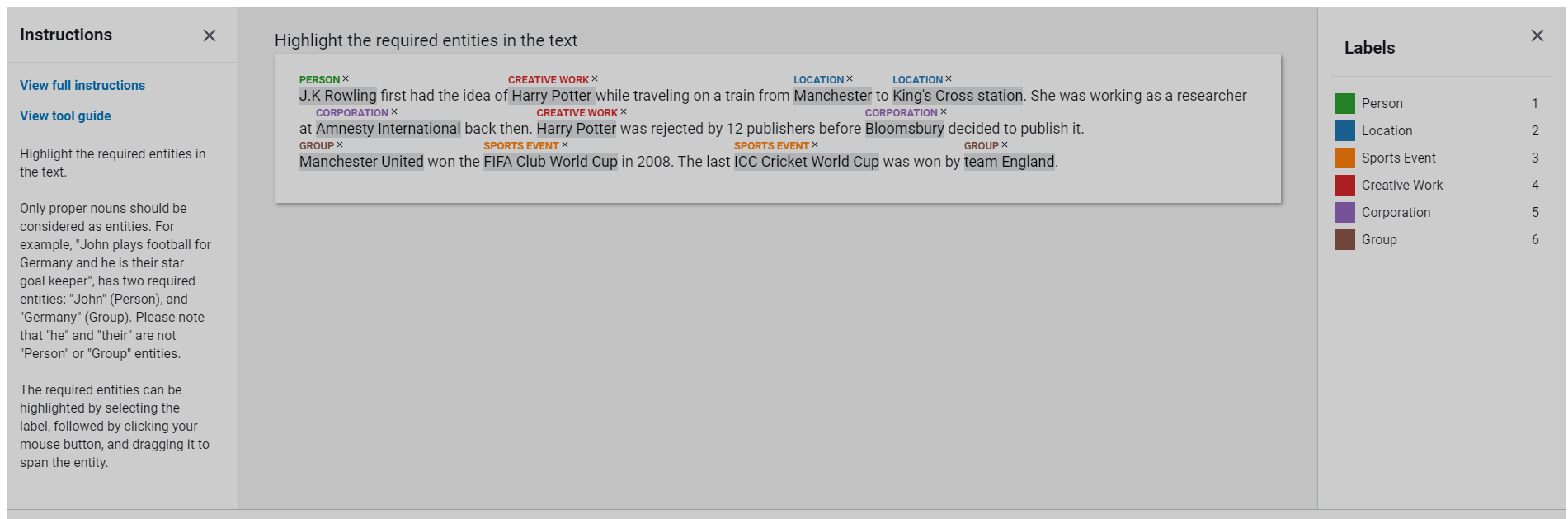}
\centering
\caption{Incomplete Instructions Issue; Not every category has an example or provided definition}
\end{figure}

\begin{figure}[h]
\includegraphics[scale=0.25]{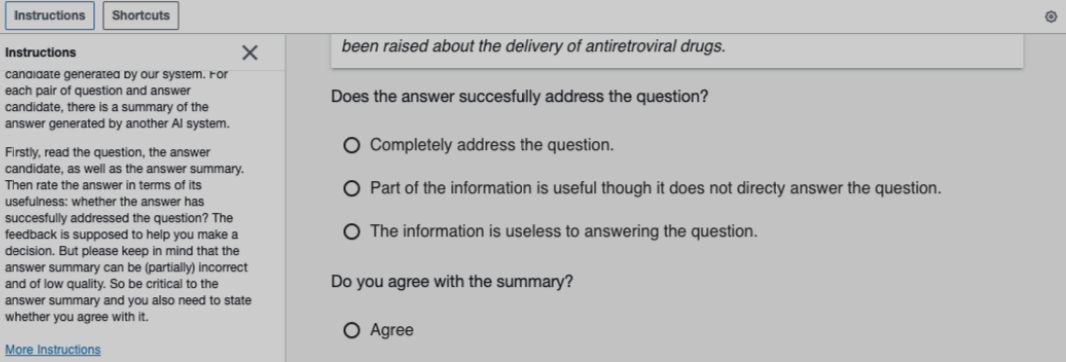}
\centering
\caption{Ambiguous/Vague Instructions Issue; The instructions and the actual multiple choice question do not match up, so it may be confusing to the worker}
\end{figure}

\begin{figure}[h]
\includegraphics[scale=0.10]{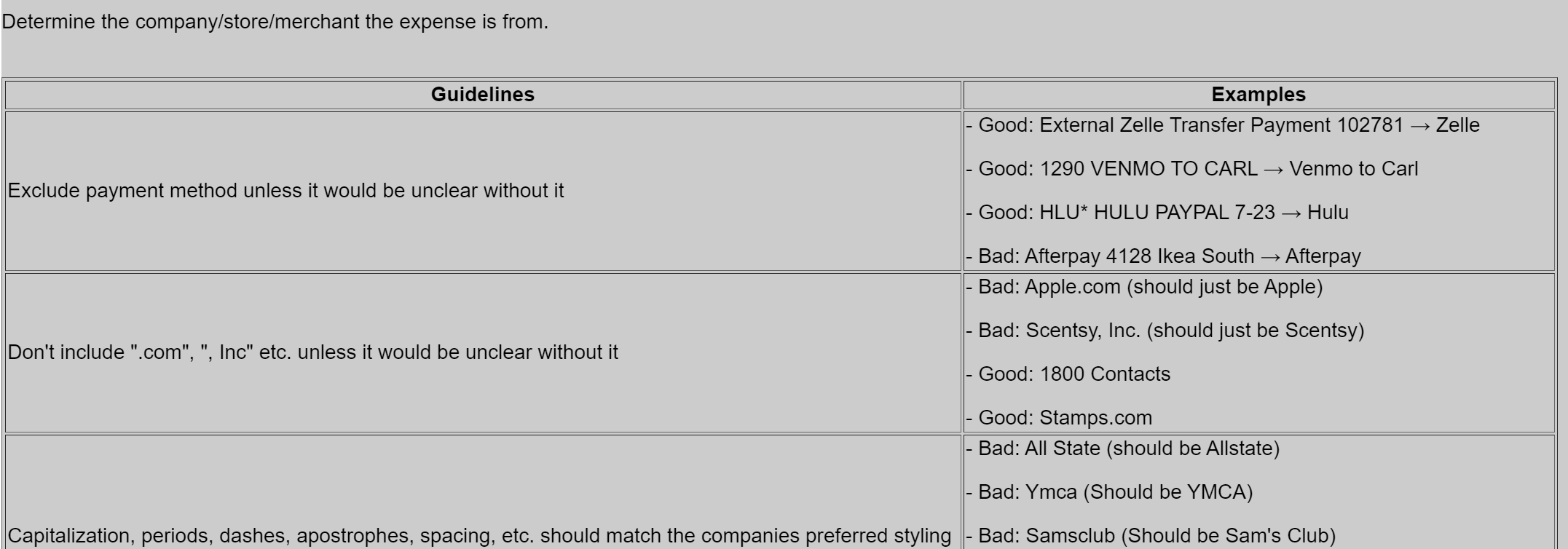}
\centering
\caption{Unrelated Instructions Issue; The instructions seem clear and straightforward. However, as seen in the image below, the task itself is still very confusing.}
\end{figure}

\begin{figure}[h]
\includegraphics[scale=0.15]{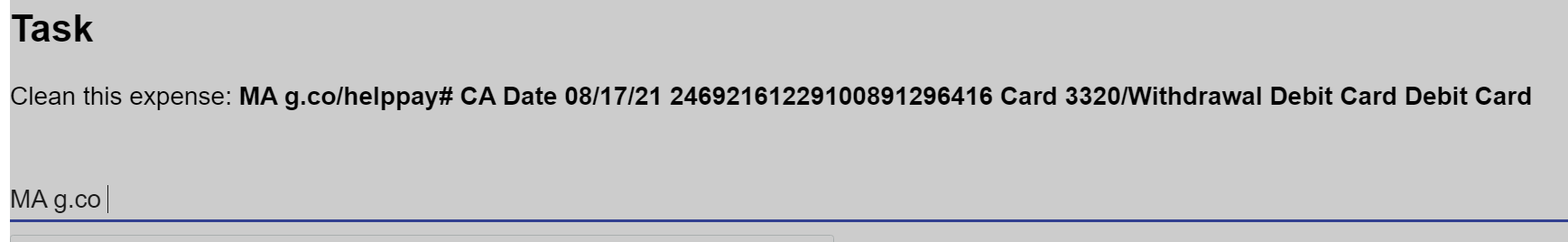}
\centering
\caption{Unrelated Instructions Issue; The task is confusing even with the instructions.}
\end{figure}

\section{TurkerView Metrics}
\begin{figure}[h]
\includegraphics[scale=0.75]{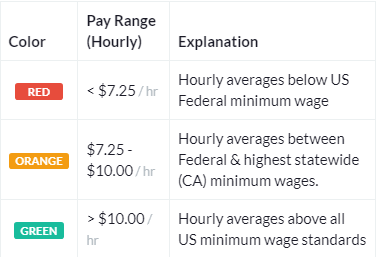}
\centering
\caption{Wage Assessment of Requesters by Workers}
\end{figure}

\begin{figure}[h]
\includegraphics[scale=0.75]{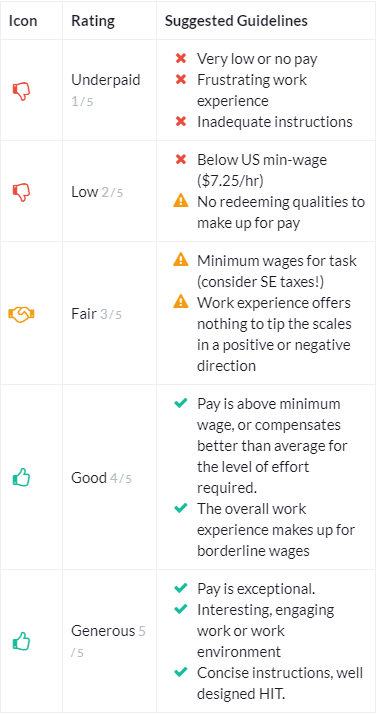}
\centering
\caption{Payment Assessment of Requesters and HITs by Workers}
\end{figure}

\begin{figure}[h]
\includegraphics[scale=0.75]{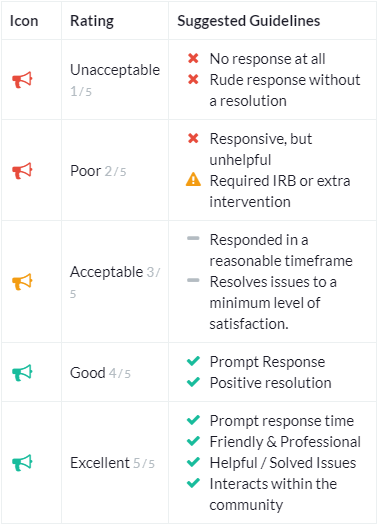}
\centering
\caption{Communication Assessment of Requesters and HITs by Workers}
\end{figure}

\begin{figure}[h]
\includegraphics[scale=0.75]{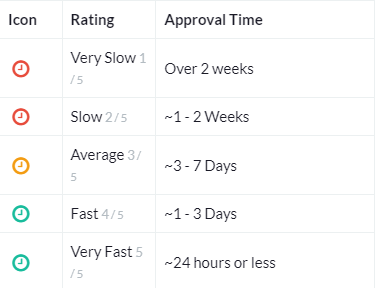}
\centering
\caption{Approval Time Assessment of Requesters and HITs by Workers}
\end{figure}

\end{document}